\pdfoutput=1

\documentclass[11pt]{article}

\usepackage[preprint]{acl}
\usepackage[T1]{fontenc}
\usepackage[utf8]{inputenc}
\usepackage{times}
\usepackage{latexsym}
\usepackage{microtype}
\usepackage{inconsolata}

\usepackage{CJKutf8}
\usepackage{bm}
\usepackage{amsmath}
\usepackage{covington}
\usepackage{booktabs}
\usepackage{siunitx}
\usepackage{newtxmath}
\usepackage{graphicx}

\usepackage{tipa}
\robustify\bfseries
\usepackage{tikz}
\usetikzlibrary{arrows.meta}
\usepackage{pgfplots} \pgfplotsset{compat=1.18}

\title{Language Complexity and Speech Recognition Accuracy: \\ Orthographic Complexity Hurts, Phonological Complexity Doesn't}

\author{Chihiro Taguchi \\
  University of Notre Dame \\
  \texttt{ctaguchi@nd.edu} \\\And
  David Chiang \\
  University of Notre Dame \\
  \texttt{dchiang@nd.edu} \\}

\begin{document}
\maketitle
\begin{abstract}
We investigate what linguistic factors affect the performance of Automatic Speech Recognition (ASR) models.
We hypothesize that orthographic and phonological complexities both degrade accuracy.
To examine this, we fine-tune the multilingual self-supervised pretrained model Wav2Vec2-XLSR-53 on 25 languages with 15 writing systems, and we compare their ASR accuracy, number of graphemes, unigram grapheme entropy, logographicity (how much word/morpheme-level information is encoded in the writing system), and number of phonemes.
The results demonstrate that orthographic complexities significantly correlate with low ASR accuracy, while phonological complexity shows no significant correlation.
\end{abstract}
\section{Introduction}
When a human learns a second language, a complex writing system and a complex phonological system can both be obstacles to language learning.
For example, learners of a language like Chinese may spend years learning thousands of characters, while Japanese learners of English commonly struggle with mastering the two liquid phonemes /r/ and /l/ that are not distinguished in Japanese.
By analogy with these observations, we may ask: do computational models of language, like humans, also struggle with these linguistic complexities?
In this paper, we investigate the relationship between the accuracy of Automatic Speech Recognition (ASR) and linguistic complexity, specifically orthographic and phonological complexity.

To answer this question, this study proposes three hypotheses about factors that may make learning ASR hard. 
\paragraph{Hypothesis 1.} 
If a language has more character (\emph{grapheme}) types, then ASR accuracy gets lower.
This idea corresponds to the example of Chinese mentioned above.

\paragraph{Hypothesis 2.}
The more a language's writing system encodes word- or morpheme-level information, the more ASR accuracy decreases. \Citet{sproat-gutkin-2021-taxonomy} call this property \emph{logographicity}, as opposed to \emph{phonographicity}, which means that a language's written form is more predictable from its spoken form.

\paragraph{Hypothesis 3.}
If a language has more sound (\emph{phoneme}) types, then ASR accuracy gets lower.
This idea corresponds to the example of Japanese learners of English mentioned above.

To test these hypotheses, we fine-tune the same pre-trained ASR model (Wav2Vec2-XLSR-53) on 25 languages with 15 different orthographies.

The results demonstrate a significant correlation of ASR accuracy with measures related to orthographic complexity, while no significant correlation is observed with phonological complexity.
\section{Related Work}\label{sec:related}
\paragraph{Multilingual ASR.}
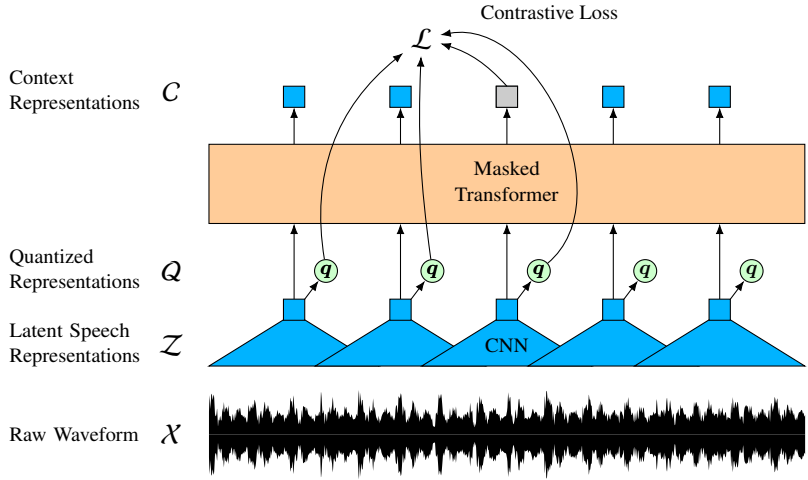
\begin{figure*}[t]
    \centering
    \begin{center}
\scalebox{0.7}{
    \begin{tikzpicture}[
        declare function={
          excitation(\t,\w) = sin(\t*\w);
          noise = 2*rnd;
          source(\t) = excitation(\t,20) + noise;
          filter(\t) = 1 - abs(sin(mod(\t, 50)));
          speech(\t) = source(\t)*filter(\t);
        }
        ,every path/.style={line width=0.02cm}
        ]

        \pgfmathsetmacro{\randseed}{42};

        \colorlet{CNNcol}{blue!30!cyan};
        \colorlet{TRNcol}{orange!40};
        \colorlet{qcol}{green!20};

        \pgfmathsetseed{\randseed};
        \fill[black, x=0.014cm, y=0.2cm] (0,0) -- plot [domain=0:800, samples=800,smooth] (\x,{1.5+speech(\x)}) -- (800,0) -- cycle;
        \pgfmathsetseed{\randseed};
        \fill[black, x=0.014cm, y=0.2cm] (0,0) -- plot [domain=0:800, samples=800, smooth] (\x,{-1.5-speech(\x)}) -- (800,0) -- cycle;

        \coordinate (T1) at (0,1.3);
        \coordinate (T2) at (3.2,1.3);
        \coordinate (T3) at (1.6,2.3);

        \coordinate (S1) at (1.4,2.56666666667);
        \coordinate (S2) at (1.8,2.56666666667);
        \coordinate (S3) at (1.8,2.166666666667);
        \coordinate (S4) at (1.4,2.166666666667);
        \coordinate (Scorner) at (1.8,2.566666666667);

        \coordinate (N1) at (1.6,2.566666666667);
        \coordinate (N2) at (1.6,4);
        \draw[-{Latex[length=0.2cm]}] (N1) -- (N2);

        \coordinate (N3) at (1.6,5.5);
        \coordinate (N4) at (1.6,6.2);
        \draw[-{Latex[length=0.2cm]}] (N3) -- (N4);
        \coordinate (Circ) at (2.2,3.1);

        \filldraw[fill=CNNcol] (T1) -- (T2) -- (T3) -- cycle;
        \filldraw[fill=CNNcol] (S1) -- (S2) -- (S3) -- (S4) -- cycle;
        \node[draw, circle, fill=qcol, inner sep=1pt] (circle) at (Circ) {$\bm{q}$};
        \draw[-{Latex[length=0.2cm]}] (Scorner) -- (circle);

        \filldraw[fill=CNNcol] ([yshift=4.03333333333cm] S1) -- ([yshift=4.03333333333cm] S2) -- ([yshift=4.03333333333cm] S3) -- ([yshift=4.03333333333cm] S4) -- cycle;
        
        \foreach \i in {1,2,3,4} {
            \draw[-{Latex[length=0.2cm]}] ([xshift=2cm*\i] N1) -- ([xshift=2cm*\i] N2);
            \draw[-{Latex[length=0.2cm]}] ([xshift=2cm*\i] N3) -- ([xshift=2cm*\i] N4);

            \filldraw[fill=CNNcol] ([xshift=2cm*\i] T1) -- ([xshift=2cm*\i] T2) -- ([xshift=2cm*\i] T3) -- cycle;

            \filldraw[fill=CNNcol] ([xshift=2cm*\i] S1) -- ([xshift=2cm*\i] S2) -- ([xshift=2cm*\i] S3) -- ([xshift=2cm*\i] S4) -- cycle;

            \filldraw[fill=CNNcol] ([xshift=2cm*\i, yshift=4.03333333333cm] S1) -- ([xshift=2cm*\i, yshift=4.03333333333cm] S2) -- ([xshift=2cm*\i, yshift=4.03333333333cm] S3) -- ([xshift=2cm*\i, yshift=4.03333333333cm] S4) -- cycle;

            \node[draw, circle, fill=qcol, inner sep=1pt] (circle) at ([xshift=2cm*\i] Circ) {$\bm{q}$};
            \draw[-{Latex[length=0.2cm]}] ([xshift=2cm*\i] Scorner) -- ([xshift=2cm*\i] circle);

        }
        
        \filldraw[fill=TRNcol] (0,4) rectangle (11.2,5.5);

        \filldraw[fill=gray!40] ([xshift=2cm*2, yshift=4.03333333333cm] S1) -- ([xshift=2cm*2, yshift=4.03333333333cm] S2) -- ([xshift=2cm*2, yshift=4.03333333333cm] S3) -- ([xshift=2cm*2, yshift=4.03333333333cm] S4) -- cycle;

        \node[text width = 2.5cm, align=center] (trn) at (5.6, 4.8) {Masked\\ Transformer};
        \node[] (cnn) at (5.6, 1.7) {CNN};

        \node[text width = 2.5cm] (wvf) at (-2.5, 0) {Raw Waveform };
        \node[] (X) at (-0.7, 0) {\Large $\mathcal{X}$};

        \node[text width = 2.5cm] (ltn) at (-2.5, 1.7) {Latent Speech Representations};
        \node[] (Z) at (-0.7, 1.7) {\Large $\mathcal{Z}$};

        \node[text width = 2.5cm] (qnt) at (-2.5, 3.1) {Quantized\\ Representations};
        \node[] (Q) at (-0.7, 3.1) {\Large $\mathcal{Q}$};
        
        \node[text width = 2.5cm] (ctx) at (-2.5, 6.5) {Context\\ Representations};
        \node[] (C) at (-0.7, 6.5) {\Large $\mathcal{C}$};

        \node[] (ctr) at (6.4, 8) {Contrastive Loss};
        \node[] (L) at (4, 7.5) {\Large $\mathcal{L}$};


        \node[circle, inner sep=1pt] (circle) at (Circ) {$\bm{q}$};
        \draw[-{Latex[length=0.2cm]}] (circle) to[bend left=30]  (L);
        \node[circle, inner sep=1pt] (circle) at ([xshift=2cm] Circ) {$\bm{q}$};
        \draw[-{Latex[length=0.2cm]}] ([xshift=2cm] circle) to[bend left=5] (L);
        \node[circle, inner sep=1pt] (circle) at ([xshift=4cm] Circ) {$\bm{q}$};
        \draw[-{Latex[length=0.2cm]}] ([xshift=4cm] circle) to[bend right=75] (L);
        \draw[-{Latex[length=0.2cm]}] ([xshift=2cm*2+0.2cm, yshift=4.03333333333cm] S1) to[bend right=15] (L);

    \end{tikzpicture}
}
\end{center}
    \caption{Visualization of the self-supervised pretraining step of Wav2Vec 2.0.
    }
    \label{fig:wav2vec2}
\end{figure*}

With the successful development of Transformer-based architectures, the field of multilingual ASR has also achieved drastic improvements.
Wav2Vec 2.0 \citep{baevski2020wav2vec} is a framework for learning speech representations with a self-supervised method like BERT \citep{devlin2019bert}.
For each span of speech with a fixed length, this architecture first obtains a latent representation $\bm{z}$ through a feature encoder with a CNN, and computes its discretized vector $\bm{q}$ via product quantization \citep{jegou2011product}. Next, it masks some $\bm{z}$ with a certain probability; the objective of this self-supervised training is then to predict the quantized vector $\bm{q}$ for the masked representation.
The pretraining step is illustrated in Figure \ref{fig:wav2vec2}.

Wav2Vec 2.0 is known to perform well for various speech recognition tasks by fine-tuning the pre-trained model.
Wav2Vec2-XLSR-53 \citep{conneau2020unsupervised}, which this study employs, is a 300M-parameter model pre-trained on speech samples of 53 languages.
It performs well for various languages, including those unseen in the pre-training step, by fine-tuning with a small amount of annotated data.
Since the appearance of Wav2Vec2.0, other larger multilingual Wav2Vec 2.0 pretrained models have been developed, such as Wav2Vec2-XLS-R \citep{babu2021xlsr} which is pretrained on 0.5M hours from 128 languages and Wav2Vec2-BERT \citep{communication2023seamless} pretrained on 4.5M hours from 143 languages.

The current state-of-the-art in ASR for some languages has been achieved by Whisper \citep{radford2022robust}, which is an encoder-decoder architecture based on next token prediction with weak supervision.
However, unlike Wav2Vec 2.0, a pretrained Whisper model is weakly supervised; namely, the pretraining data contains some labeled data.
Importantly, fine-tuning on the languages that have labeled data in Whisper's pretraining step is known to remarkably boost the performance, while there is less observed improvement between the languages included in Wav2Vec 2.0 pretraining and those not included \cite{rouditchenko23_interspeech}.
To avoid the performance difference biased by labeled pretraining data, our work conducts experiments on Wav2Vec2-XLSR-53 through fine-tuning.

\paragraph{Logographicity.}
In some languages, the spelling of a token encodes word- or morpheme-level information that is not predictable from the token's pronunciation alone.
For instance, English /ra\textipa{I}t/ can be spelled as <write>, <right>, <rite>, or <wright>, which all have different meanings.
\Citet{sproat-gutkin-2021-taxonomy} calls this property \textit{logographicity}.
To measure logographicity, they train a phoneme-to-grapheme model and look at how widely dispersed its attention is to see how context-dependent the orthography is (see Section~\ref{sec:h2} for details).
\section{Experimental Setup and Methods}
This section describes the setup and the methods used in the experiments.\footnote{The code used in the experiments is available at: \url{https://github.com/ctaguchi/ASRcomplexity}}

\subsection{Dataset}
In this experiment, we use Common Voice 16.1 \citep{ardila-etal-2020-common} for every language examined except English and Korean.
We use LibriSpeech \cite{7178964} for English instead because Common Voice English contains a number of non-native speech samples and Zeroth-Korean\footnote{\url{https://github.com/goodatlas/zeroth}} 
because the Korean subset in Common Voice 16.1 does not have enough samples.
Since LibriSpeech and Zeroth-Korean have a longer maximum audio length than Common Voice 16.1, long audio samples are filtered out.
For each setting, we keep extracting training data samples until the total sample length reaches 10,000 seconds.
In doing so, we aim to standardize the training dataset size across the settings rather than to rely on number of samples that vary in their audio length.

\subsection{Pre-trained model and fine-tuning}
We use Wav2Vec2-XLSR-53\footnote{\url{https://huggingface.co/facebook/wav2vec2-large-xlsr-53}} as the base pre-trained model for every setting and fine-tune it for each target language and writing system.
The fine-tuning step adds supervised training to the pre-trained model with Connectionist Temporal Classification (CTC, \citealp{graves-etal-2006-ctc})
as illustrated in Figure \ref{fig:wav2vec2}.
In the experiment, the same hyperparameters were used for every fine-tuned model; among others, we ran $20$ epochs and the learning rate was set to $0.0003$.
Each experiment took approximately two hours on two GPU cores (NVIDIA A10).
Also, punctuation was removed at the preprocessing step.\footnote{The training code is available in the repository: \url{https://github.com/ctaguchi/ASRcomplexity}}

\subsection{Graphemes}
Our first hypothesis is that a language's higher number of graphemes worsens ASR accuracy.
To test this in a controlled way, we include among our fine-tuning settings some languages that have multiple scripts, namely, Japanese, Korean, and Chinese.

For Japanese, we use the following three systems: a combination of \textit{Kanji} (Chinese characters) and \textit{Kana} (syllabary), which is the default orthography, \textit{Kana}-only, and \textit{Romaji}-only (romanized Kana).
For example, the word for ``the Japanese language'' in Japanese is <\begin{CJK*}{UTF8}{ipxm}日本語\end{CJK*}>, and can be transliterated as <\begin{CJK*}{UTF8}{ipxm}ニホンゴ\end{CJK*}> in \textit{Kana} and as <nihongo> in \textit{Romaji}.
\textit{Kana} can be uniquely mapped to \textit{Romaji} but not vice versa.
The tokenization and conversion from the default orthography into \textit{Kana}-only is done by SudachiPy \citep{takaoka-etal-2018-sudachi}.
Then, these \textit{Kana} are romanized with the \texttt{pykakasi} library.\footnote{\url{https://pykakasi.readthedocs.io}}

For Korean, we use \textit{Hangul syllables}, a syllabary writing system where each syllable character is composed of phonemic components, and \textit{Hangul Jamo}, which is decomposed Hangul so that each character represents a phoneme.
For example, /han \textipa{g}\textturnm l/ is written as \begin{CJK*}{UTF8}{mj}한글\end{CJK*} in the default orthography (\textit{Hangul syllables}) and can be decomposed into six \textit{Hangul Jamo} letters:
<\begin{CJK*}{UTF8}{mj}ㅎ\end{CJK*}> /h/,
<\begin{CJK*}{UTF8}{mj}ㅏ\end{CJK*}> /a/,
<\begin{CJK*}{UTF8}{mj}ㄴ\end{CJK*}> /n/,
<\begin{CJK*}{UTF8}{mj}ㄱ\end{CJK*}> /\textipa{g}/,
<\begin{CJK*}{UTF8}{mj}ㅡ\end{CJK*}> /\textturnm/, and
<\begin{CJK*}{UTF8}{mj}ㄹ\end{CJK*}> /l/.
In the preprocessing step, \textit{Hangul syllables} are converted to \textit{Hangul Jamo} by the \texttt{g2pK} library \cite{park2019g2pk}.

For Chinese, three writing systems are used: 
\textit{Hanzi} (Chinese characters),
\textit{Zhuyin} (semi-syllabary),
and \textit{Pinyin} (romanized).
For example, the Chinese word for ``the Chinese language'' is <\begin{CJK*}{UTF8}{bsmi}漢語\end{CJK*}> in \emph{Hanzi}, and can be expressed as <\begin{CJK*}{UTF8}{bsmi}\mbox{ㄏㄢ$^{\text{ˋ}}$} \mbox{ㄩˇ}\end{CJK*}> in \textit{Zhuyin} and as <hànyǔ> in \textit{Pinyin}.
\textit{Zhuyin} and \textit{Pinyin} can be converted to each other by rules.
In our implementation, we convert \textit{Hanzi} into \textit{Zhuyin} and \textit{Pinyin} using the \texttt{dragonmapper} library.\footnote{\url{https://github.com/tsroten/dragonmapper}}

To measure the impact of the grapheme size to ASR accuracy, we employ two metrics.
One is to naively count all the character (grapheme) types in the training data.
The other is to calculate the unigram character entropy of the training data, to capture the fact that not all character types appear with the same probability.
In fact, it is known that Chinese \emph{Hanzi} have a Zipfian distribution \cite{deng-et-al-2014-rank}.
The unigram character entropy is computed as $H(C) = -\sum_{c \in \mathcal{C}} p(c) \log p(c)$, where $\mathcal{C}$ is the set of character types in the corpus.

\begin{figure*}
    \centering
    \includegraphics[width=.8\linewidth]{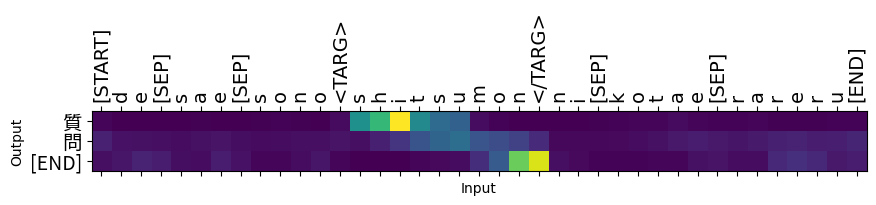}
    \includegraphics[width=.8\linewidth]{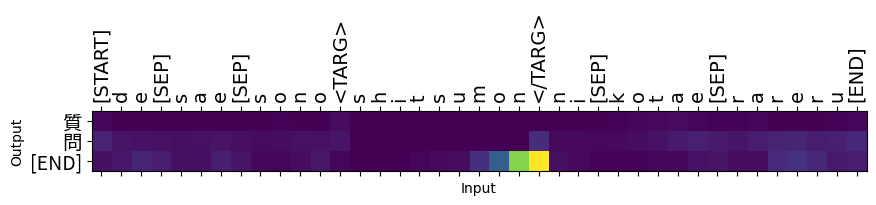}
    \caption{A visualization of attention masking.
    The top matrix shows the original distributions of attention scores with a Japanese phoneme input and an orthographic output of the target word.
    The bottom matrix has zeroed-out attention values for the cells corresponding to the target word.
    The logographicity score $S_\text{token}$ measures how much information is retained after masking.
    Values near 1 are in yellow and those near 0 in dark purple.
    }
    \label{fig:mask}
\end{figure*}

\subsection{Logographicity}\label{sec:h2}

Hypothesis 2 claims that the more logographic a language is (that is, the more irregular the mapping from pronunciation to orthography is in a language), the harder it is for an ASR model to transcribe the language.
To measure logographicity, we follow \citet{sproat-gutkin-2021-taxonomy} and train a model that predicts a word's orthography given its phonemes and context.
If a language is phonographic (\textit{i.e.}, a word's orthography can be easily reconstructed by how it is pronounced), then the attention matrix of a learned model would only attend to a word's phonemes and would not attend its surrounding context.
On the other hand, if a language is logographic and a word's pronunciation may depend on the context, then the model would attend to other surrounding words.

This attention-based metric of logographicity is calculated as follows.
Given an attention matrix $A$ and a mask matrix $M$, $M \circ A$ is their component-wise (Hadamard) product.
The mask matrix $M$ is a matrix of the same size as $A$ whose entries $i,j$ are 0 if $0 \leq i < k$ where $k$ is the length of the target word's pronunciation and $m \leq j \leq n$ where $m$ is the left edge of the target word in a text and $n$ the right edge.
Then, the attention spread $S_w$ for a word $w$ is:
$$S_w= \frac{\sum_{i,j} (M \circ A)_{i,j}}{\sum_{i,j}A_{i,j}}.$$
To apply this to a writing system of a language, one can compute the average attention spread of a word over a corpus $\mathcal{D}$:
$$S_\text{token}=\frac{\sum_{w \in \mathcal{D}} S_w}{|\mathcal{D}|}.$$
A phonographic language would have lower $S_\text{token}$ than a logographic language.
See Figure \ref{fig:mask} for an example of an attention matrix and a masked attention matrix.

\begin{table*}[t]
    \centering
    \begin{tabular}{lll} \toprule
        Language & Script type & Dataset \\ \midrule
        Chinese & logographic & LCCC \cite{wang2020chinese} \\
        Japanese & logographic/syllabary & SNOW \cite{maruyama-yamamoto-2018-simplified}  \\
        Korean & syllabary & Korean Parallel Corpora \cite{park-etal-2016-korean} \\
        Thai & abugida & ThaiGov V2 Corpus \cite{phatthiyaphaibun-etal-2023-pythainlp} \\
        Arabic & abjad & Rasaif Classical-Arabic--English Parallel Texts \\
        English & alphabetic & Europarl \cite{koehn-2005-europarl} \\
        French & alphabetic & Europarl \\
        Italian & alphabetic & Europarl \\
        Czech & alphabetic & Europarl \\
        Swedish & alphabetic & Europarl \\
        Dutch & alphabetic & Europarl \\
        German & alphabetic & Europarl \\
        \bottomrule
    \end{tabular}
    \caption{Details of the datasets for measuring logographicity.
    The original Arabic data were published on \url{https://rasaif.com}.}
    \label{tab:logo_datasets}
\end{table*}

In this experiment, we measure the logographicity of various languages with different types of writing systems.
See Table \ref{tab:logo_datasets} for the list of languages covered and the datasets used.
For each language, we extracted 10k samples and used 80\% as the training set and 20\% as the validation set.
Since our objective is not to measure the accuracy but to measure the attention spread, we do not prepare a test set.
Punctuation is removed, and samples with numerical symbols are omitted.
The phoneme string contains seven tokens, with the phonemized target token in the middle and three tokens on its left- and right-hand sides.
In the phoneme string, the target word is enclosed by tag tokens \texttt{<TARG>} and \texttt{</TARG>}, and word boundaries are marked by \texttt{[SEP]}.
Special characters \texttt{[START]} and \texttt{[END]} are added to the beginning and the end of both input and output, respectively.

For phonemization, we used the same converters for Japanese, Korean, and Chinese.
For Thai romanization, we used PyThaiNLP's transliteration module \cite{phatthiyaphaibun-etal-2023-pythainlp}.
For English phonemization, we used the CMU Pronouncing Dictionary\footnote{\url{http://www.speech.cs.cmu.edu/cgi-bin/cmudict}} via \texttt{eng-to-ipa} package.\footnote{\url{https://pypi.org/project/eng-to-ipa}}
Other alphabetic languages and Arabic were converted into IPA 
by \texttt{gruut} library.\footnote{\url{https://github.com/rhasspy/gruut}}

The architecture is an attention-based sequence-to-sequence encoder--decoder model.
The encoder consists of an embedding layer and a bidirectional GRU layer.
The decoder consists of an embedding layer and a unidirectional GRU layer, followed by a single cross-attention layer that receives context vectors from the encoder.
The attention matrix that we focus on here is the result of multi-head attention before residual connection and layer normalization.
The hyperparameters are the same as the implementation by \citet{sproat-gutkin-2021-taxonomy}, except that we run 10 epochs with a batch size of~64.


\subsection{ASR Metrics}
To measure ASR accuracy, we employ two approaches in this experiment: Character Error Rate (CER) and Calibrated Errors Per Second (CEPS).
The following describes the details of these metrics.

\subsubsection{Character Error Rate}
CER is calculated as:
$$\text{CER} = \dfrac{S + D + I}{N}$$
where $S$, $D$, $I$ are the numbers of substitution, deletion, and insertion errors, respectively, and $N$ is the length of the reference text.
Though Word Error Rate (WER) is also a widely used metric in the ASR literature, we do not use it in this experiment because the notion of \textit{word} or \textit{token} varies across languages.
For example, Tatar, an agglutinative language, can take multiple inflectional suffixes to form one token as in (\ref{tatar}), where the English translation is expressed with seven tokens.
\begin{covexample}
    \label{tatar}
    \digloss
    {tan{\i}-{\c s}-qan-{\i}b{\i}z-{\u g}a}
    {get.to.know-each.other-participle-our-to}
    {to our getting to know each other}
\end{covexample}
The use of WER can unfairly harm the evaluation of such agglutinative languages compared to analytic languages.
In addition, some languages and writing systems are less clear as to what a word is, particularly those written without a whitespace character (Japanese, Chinese, and Thai in our experiments).
In contrast, the notion of a (Unicode) character is always clear, so we use CER using Unicode characters.

\subsubsection{Errors Per Second}
A problem with ASR evaluation metrics that are solely based on text like WER and CER is that they are not comparable across languages or scripts.
Consider, for example, evaluating the same system on the same data, but represented in two ways: one where the characters are \textit{Hangul Jamo} and one where the characters are \textit{Hangul syllables}.
The two settings should have the same accuracy, but will have different CERs.

We make the following simplifying assumptions.
First, all languages communicate the same amount of information per second \citep{coupe-2019} and therefore times are comparable across languages.
Second, speech is divided into equal-length slices of $\tau$ seconds each.
Third, an ASR error is an event that occurs at a single point in time, and errors are Poisson-distributed, so the probability that a given slice has $k$ errors is
\begin{equation*}
P(k; \lambda) = \frac{(\lambda \tau)^k e^{-\lambda \tau}}{k!}.
\end{equation*}

The parameter $\lambda$ can be thought of as the number of errors per second, which we propose to use as an error rate that is comparable across different segmentations, writing systems, and languages. We now describe how to estimate $\lambda$ from a run of an ASR system on test data.

First, define a slice to be a character, a word, or something else, and consider the ASR output and reference transcription as strings of slices. Let $\tau$ be the average length of a slice, in seconds, and let $n$ be the number of slices in the reference. Compute the Levenshtein distance $d$ between the output and the reference, and let $p = d/n$ be the usual normalized Levenshtein distance.

Recall that we assumed that an error occurs at a single point in time, so a single slice could contain more than one error. However, we can only detect whether a slice has at least one error; we can't distinguish between a slice with one error versus a slice with two errors.
The probabilities of a slice having no errors and at least one error are
\begin{align*}
P_\lambda(k=0) &= e^{-\lambda \tau} \\
P_\lambda(k>0) &= 1-e^{-\lambda \tau}
\end{align*}
so the log-likelihood is
\begin{align*}
L(\lambda) 
&= pn \log (1-e^{-\lambda\tau}) - (1-p)n \lambda\tau
\end{align*}
and the maximum-likelihood estimate of $\lambda$ is
\begin{align}
\lambda &= \frac1\tau \log \frac{1}{1-p}. \label{eq:ceps}
\end{align}
We call this the \emph{calibrated errors per second (CEPS)}.
Note that for $p\ll1$, this reduces to $\lambda \approx \frac{p}{\tau}$, the raw number of errors per second, but for larger $p$, the CEPS grows faster than the raw errors per second, accounting for the fact that slices may have more than one error.

\begin{figure}
\centering
\begin{tikzpicture}
\begin{axis}[xlabel={$p$},xmin=0,xmax=1,xtick={0,1},ymin=0,ymax=1,ytick={0,1},axis equal image,width=5cm]
\addplot[domain=0:1,dashed] { x };
\addplot[domain=0:1,samples=100] { ln (1/(1-x)) };
\end{axis}
\end{tikzpicture}
\caption{Calibrated errors per second (solid) compared with raw errors per second (dashed), assuming $\tau=1$.}
\label{fig:ceps_graph}
\end{figure}
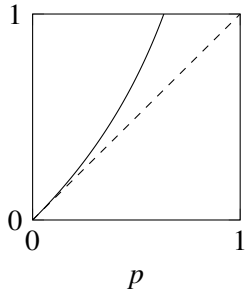
\section{Results}\label{sec:results}
This section reports the results of the experiments to test the three hypotheses.
Table \ref{tab:results} summarizes the results across languages and metrics.
Comparisons with different writing systems within Japanese, Korean, and Chinese clearly demonstrate that logographic writing systems like Japanese \textit{Kanji} and Chinese \textit{Hanzi}, as well as writing systems that are phonographic but have hundreds of syllable characters like Korean \textit{Hangul syllables}, make ASR models harder to learn to transcribe correctly.
Indeed, there is a significant correlation ($p < 0.05$) between CER and orthography-related variables (the number of graphemes, unigram entropy, and logographicity score), as shown in the correlation matrix in Table \ref{tab:corr}.
On the other hand, there is no significant correlation ($p > 0.05$) in Table~\ref{tab:corr} between ASR accuracy and the number of phoneme types.

We can also observe that CEPS mitigates the score deterioration coming from purely orthographic differences.
For instance, though \textit{Hangul syllables} and \textit{Hangul Jamo} only differ in how phoneme characters are encoded, we see a large difference in CER ($28.21$ and $16.72$, respectively).
However, in CEPS, \textit{Hangul syllables} have a slightly higher score than \textit{Hangul Jamo} ($2.63$ and $3.23$, respectively).
In addition, as Table~\ref{tab:corr} demonstrates, CEPS has less significant correlation with the orthographic factors ($|\mathcal{C}|$: $0.49$, $H(\mathcal{C})$: $0.41$, $S_\text{token}$: $0.61$) than CER does ($|\mathcal{C}|$: $0.85$, $H(\mathcal{C})$: $0.81$, $S_\text{token}$: $0.76$).

\begin{table*}
    \centering
    \begin{tabular}{llSSrSSS}\toprule
        Language & Writing system & {CER (\%)$^\downarrow$} & {CEPS} & $|\mathcal{C}|$ & {$H(\mathcal{C})$} & {$S_\text{token}$ (\%)} & $|\Phi|$ \\ \midrule
        Japanese & \textit{Kanji} + \textit{Kana} & 58.117 & 7.21125 & 1702 & 7.7374 & 44.98 & 27 \\
        & \textit{Kana} & 29.713 & 3.48092 & 92 & 5.6263 & 41.22 & 27 \\
        & \textit{Romaji} & 17.093 & 2.90943 & 27 & 3.5181 & 29.46 & 27 \\
        Korean & \textit{Hangul syllables} & 28.205 & 2.63099 & 965 & 7.9828 & 25.27 & 39.5 \\
        & \textit{Hangul Jamo} & 16.719 & 3.22982 & 62 & 4.8974 & 15.99 & 39.5 \\
        Chinese & \textit{Hanzi} & 62.814 & 2.64629 & 2155 & 9.4657 & 39.46 & 42.5 \\
        & \textit{Zhuyin} & 9.714 & 1.04249 & 49 & 4.814 & 24.32 & 42.5 \\
        & \textit{Pinyin} & 9.171 & 1.00768 & 56 & 5.0159 & 22.50 & 42.5 \\
        Thai & \textit{Thai} & 19.769 & 1.80009 & 67 & 5.244 & 20.5481 & 40.6667 \\
        Arabic & \textit{Perso--Arabic} & 40.592 & 4.78369 & 53 & 4.77 & 21.5709 & 37 \\
        English & \textit{Latin} & 3.168 & 0.57804 & 27 & 4.1719 & 19.1677 & 41.2222\\
        French & \textit{Latin} & 19.643 & 2.79086 & 69 & 4.4209 & 20.3675 & 36.75 \\
        Italian & \textit{Latin} & 14.805 & 1.84284 & 48 & 4.27 & 21.2772 & 43.3333 \\
        Czech & \textit{Latin} & 16.894 & 1.85509 & 46 & 4.9179 & 20.5717 & 39 \\
        Swedish & \textit{Latin} & 20.313 & 2.71388 & 34 & 4.5175 & 19.8070 & 35 \\
        Dutch & \textit{Latin} & 12.351 & 1.76631 & 36 & 4.196 & 19.6724 & 49.375 \\
        German & \textit{Latin} & 7.611 & 1.02797 & 48 & 4.1829 & 18.0292 & 40 \\
         \bottomrule
    \end{tabular}
    \caption{A summary of the experimental results.
    $\mathcal{C}$ and $\Phi$ are the sets of grapheme types and phoneme types, respectively, that appeared in the training data. Thus, $|\mathcal{C}|$ is the number of grapheme types, $H(\mathcal{C})$ is the unigram character entropy, $S_\text{token}$ is logographicity, and $|\Phi|$ is the number of phoneme types.
    The number of grapheme types and the unigram entropy $H(C)$ were calculated from the ASR training data.
    The number of phoneme types was retrieved from Phoible 2.0 \cite{phoible};
    when there is more than one total number of phoneme types reported, we use the averaged number.
    }
    \label{tab:results}
\end{table*}

\begin{table}
    \centering
    \setlength{\tabcolsep}{3pt}
    \begin{tabular}{lSSSSSS} \toprule
         & {CER} & {CEPS} & $|\mathcal{C}|$ & {$H(\mathcal{C})$} & {$S_\text{token}$} & $|\Phi|$ \\ \midrule
        CER & 1 & 0.765275 & 0.854049 &  0.812184 & 0.763578 & -0.369604 \\
        CEPS & ~ & 1 & 0.485239 & 0.410018 & 0.608001 & -0.659225 \\
        $|\mathcal{C}|$ & ~ & ~ & 1 & 0.932251 & 0.718030 & -0.135608 \\
        $H(C)$ & ~ & ~ & ~ & 1 & 0.665734 & -0.078552 \\
        {$S_\text{token}$} & ~ & ~ & ~ & ~ & 1 & -0.595294 \\
        \bottomrule
    \end{tabular}
    \caption{Correlation matrix of CER and other variables based on the results in Table \ref{tab:results}.
    }
    \label{tab:corr}
\end{table}

We also ran experiments on additional languages with phonographic scripts (i.e., alphabetic or abugida writing systems), for which we were unable to measure logographicity due to lack of reliable grapheme-to-phoneme tools. These results are summarized in Table \ref{tab:additional}.
In this case, we see less but moderate positive correlations of CER with orthographic complexities ($|\mathcal{C}|$, $H(\mathcal{C})$, and $S_\text{token}$) and no correlation with phonological complexity ($|\Phi|$), as summarized in Table \ref{tab:corr-alpha}.
All of the correlation coefficients with CER were small ($\leq \pm$0.20), and there is no significant correlation ($p > 0.05$) with respect to CER.

\begin{table*}
    \centering
    \begin{tabular}{llSSrSS} \toprule
        Language    & Writing system    & {CER (\%)$^\downarrow$} & {CEPS} &  $|\mathcal{C}|$ & {$H(\mathcal{C}$)} & $|\Phi|$ \\ \midrule
        Lithuanian  & \textit{Latin}    & 19.202    & 2.59534 & 39 & 4.5545 & 52.5 \\
        Polish      & \textit{Latin}    & 12.581    & 1.62626 & 40 & 4.5558 & 36 \\
        Basque      & \textit{Latin}    & 6.282     & 0.77749 & 27 & 3.8876 & 30.7100 \\
        Indonesian  & \textit{Latin}    & 24.011    & 3.36143 & 35 & 4.0445 & 31 \\
        Kabyle      &\textit{Latin}     & 31.593    & 3.02242 & 46 & 4.3006 & 30 \\
        Swahili     & \textit{Latin}    & 17.825    & 2.13756 & 33 & 4.0023 & 36.5 \\
        Hungarian   & \textit{Latin}    & 15.41     & 1.97697 & 37 & 4.518 & 52 \\
        Russian     & \textit{Cyrillic} & 14.444    & 1.98932 & 40 & 4.6458 & 39.3333 \\
        Tatar       & \textit{Cyrillic} & 21.429    & 3.27128 & 43 & 4.7247 & 43 \\
        Abkhaz      & \textit{Cyrillic} & 15.087    & 1.66124 & 41 & 4.6032 & 66 \\
        Georgian    & \textit{Georgian} & 14.69     & 1.77749 & 37 & 4.2946 & 33.75 \\
        Armenian    & \textit{Armenian} & 10.871    & 1.44695 & 49 & 4.5738 & 36.5 \\
        Hindi       & \textit{Devanagari} & 21.812  & 2.44429 & 119 & 5.0979 & 68.4 \\
        \bottomrule
    \end{tabular}
    \caption{Additional experimental results on languages with phonographic writing systems.
    }
    \label{tab:additional}
\end{table*}

\begin{table}
    \centering
    \setlength{\tabcolsep}{4pt}
    \begin{tabular}{lSSSSS} \toprule
                        & {CER} & {CEPS} & $|\mathcal{C}|$ & {$H(\mathcal{C})$} & $|\Phi|$ \\
                        \midrule
        CER             & 1 & 0.894968 & 0.197257 & 0.160556 & -0.184143 \\
        CEPS            & ~ & 1 & 0.180008 & 0.165454 & 0.019303 \\
        $|\mathcal{C}|$ & ~ & ~ & 1 & 0.717995 & 0.582645 \\
        $H(C)$          & ~ & ~ & ~ & 1 & 0.614899 \\
        \bottomrule
    \end{tabular}
    \caption{Correlation matrix of CER and other variables based on the results of phonographic languages.}
    \label{tab:corr-alpha}
\end{table}

\begin{figure}
\centering
\pgfplotsset{every axis/.style={width=8cm,height=5cm,trim axis left,trim axis right,ylabel={CER}}}%
\pgfplotsset{scatter/.style={
  only marks,mark=*,mark size=1pt,
  mark options={draw=red,fill=red},
  nodes near coords*={\footnotesize\lang},
  point meta=0, 
  visualization depends on={value \thisrow{lang} \as \lang}
  }
}
\begin{tabular}{@{}l@{}}
\begin{tikzpicture}
	\begin{axis}[xlabel={$|\mathcal{C}|$}]
	\addplot[scatter] table[x=graphemes,y=cer] {data.dat};
	\end{axis}
\end{tikzpicture}
\\
\begin{tikzpicture}
	\begin{axis}[xlabel={$H(\mathcal{C})$}]
	\addplot[scatter] table[x=entropy,y=cer] {data.dat};
	\end{axis}
\end{tikzpicture}
\\
\begin{tikzpicture}
	\begin{axis}[xlabel={$S_{\text{token}}$}]
	\addplot[scatter] table[x=logographicity,y=cer] {data.dat};
	\end{axis}
\end{tikzpicture}
\\
\begin{tikzpicture}
	\begin{axis}[xlabel={$|\Phi|$}]
	\addplot[scatter] table[x=phonemes,y=cer] {data.dat};
	\end{axis}
\end{tikzpicture}
\end{tabular}
\caption{CER versus various measures of linguistic complexity.}
\end{figure}
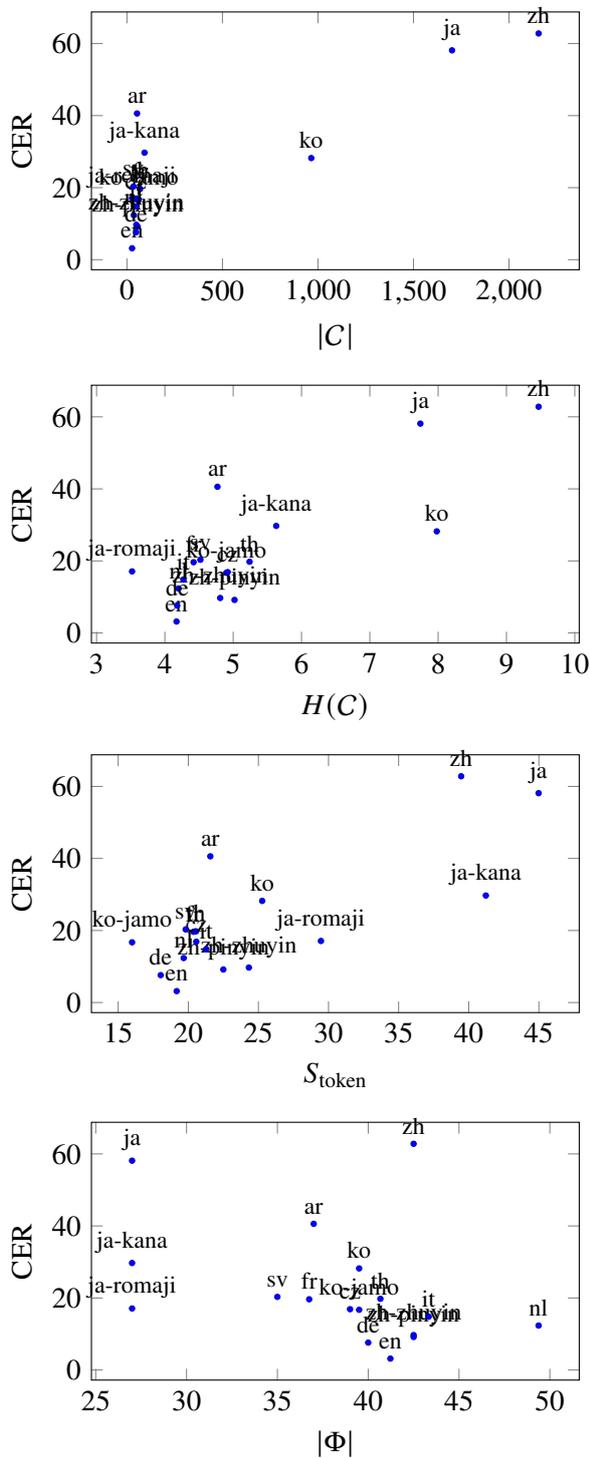

Thus, there is a clear positive correlation of ASR accuracy of fine-tuned Wav2Vec 2.0 models with orthographic complexities but not with phonological complexities.
In other words, logographic writing systems and large character inventories can harm ASR performance, supporting the first and second hypotheses, but the self-supervised pretrained model is robust to different phonological complexities, rejecting the third hypothesis.

\begin{figure}
    \centering
    \includegraphics[width=\linewidth]{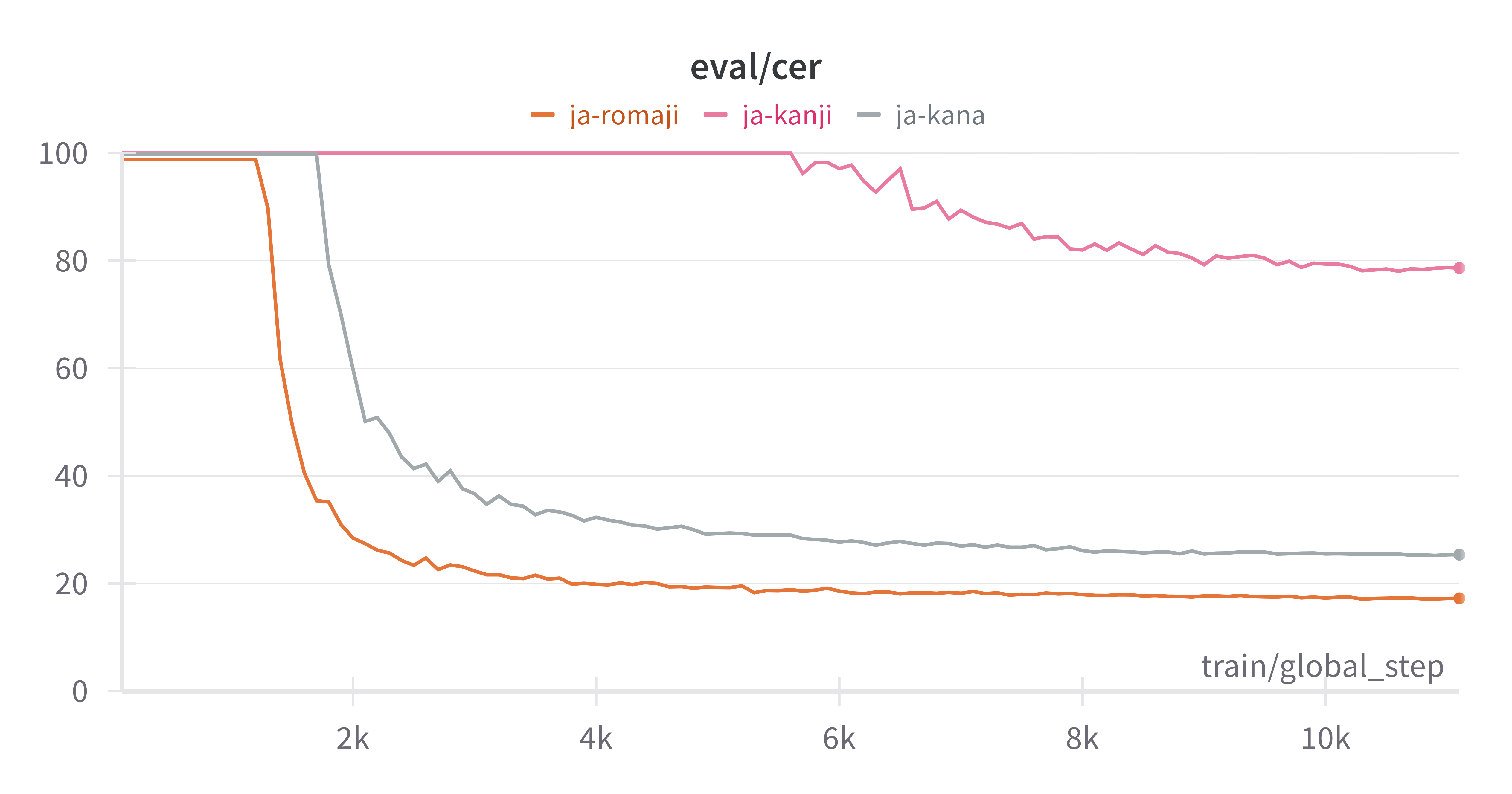}
    \includegraphics[width=\linewidth]{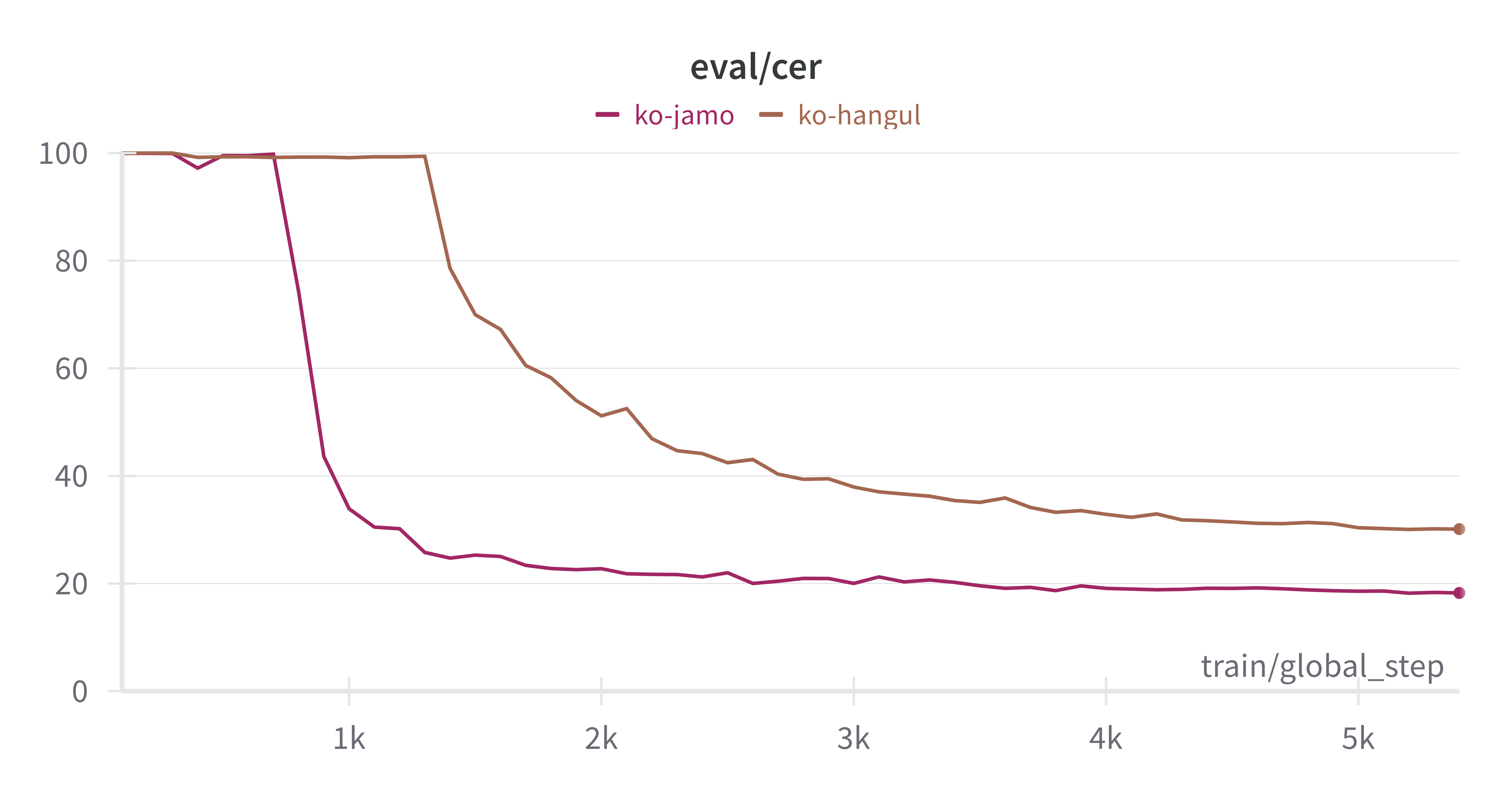}
    \includegraphics[width=\linewidth]{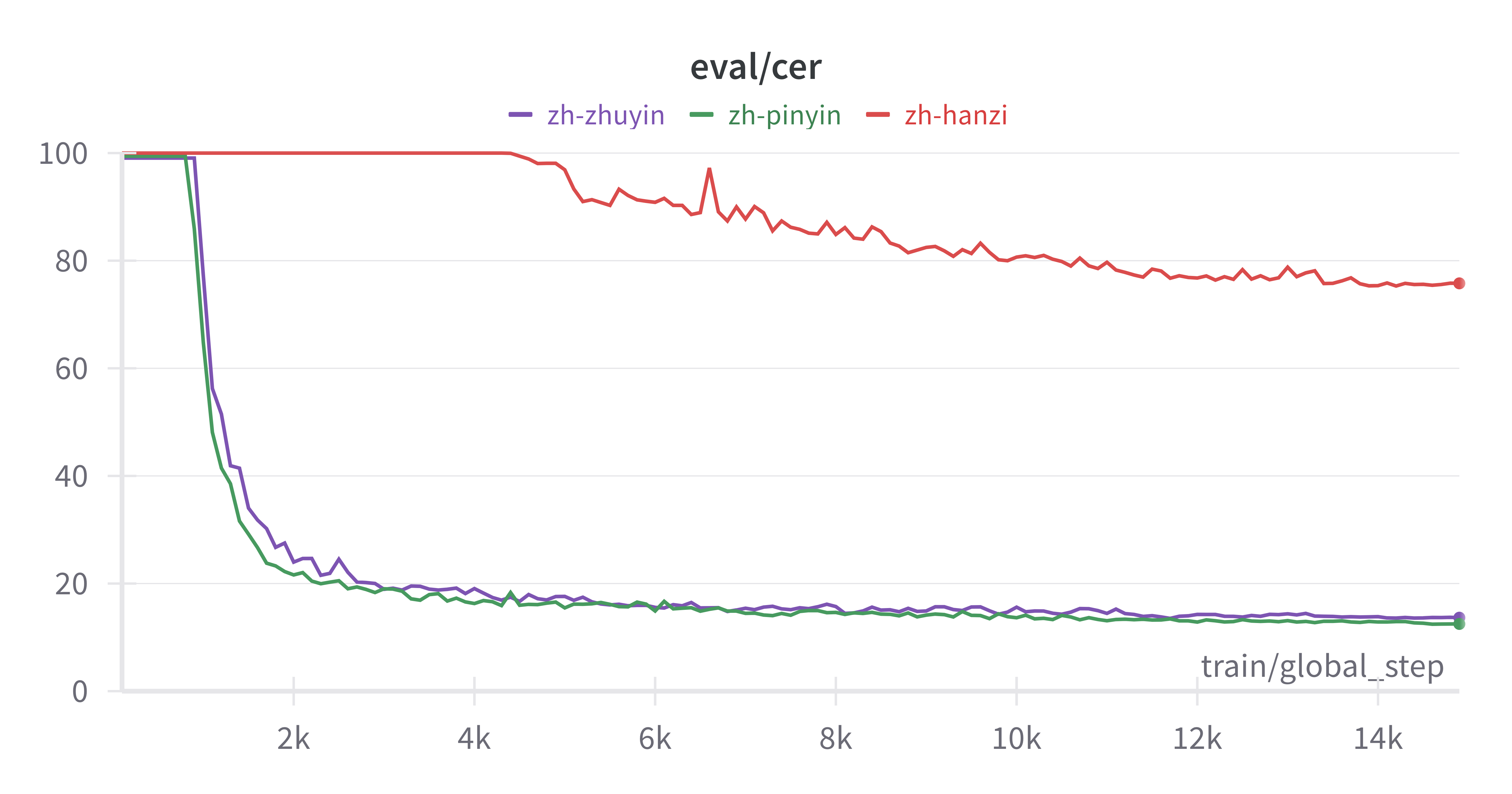}
    \caption{Comparison of validation CERs during the training with different writing systems for Japanese (top), Korean (middle), and Chinese (bottom).}
    \label{fig:eval-curve}
\end{figure}

In addition to the numerical results, there were also differences in the learning curve of fine-tuning to different writing systems.
As Figure \ref{fig:eval-curve} shows for the three writing systems of Japanese, the model struggles to learn to transcribe in a writing system with a larger inventory of graphemes.
For the mixed orthography of \textit{Kanji} and \textit{kana}, the validation CER never goes under 100\% until 5800 steps, while the \textit{Kana}-only and \textit{Romaji}-only models start to grasp transcription at much earlier steps (1800 steps and 1300 steps, respectively).
Furthermore, the curves of validation CERs over the steps is less smooth in complex orthographies (\textit{Kanji}, \textit{Hangul syllables}, and \textit{Hanzi}) than the phonographic scripts (\textit{Kana}, \textit{Romaji}, \textit{Hangul Jamo}, \textit{Zhuyin}, and \textit{Pinyin}).
This demonstrates that complex orthographies result not only in poorer ASR performance but also slower learning speed and more required training data to achieve the desired performance.
\section{Discussion}\label{sec:discussion}
In this section, we discuss the implications of the main results from the previous section.

\subsection{Low CER in English}
In Table \ref{tab:results}, one can notice that the CER of the English fine-tuned model is markedly lower than those of other languages.
There are two possible reasons for this tendency.
One is that Wav2Vec2-XLSR presumably has more English pretraining data than other languages.
The experimental results of fine-tuning Wav2Vec2-XLS-R 300M by \citet{rouditchenko23_interspeech} also show lower CER in English models than other languages.
The second possible factor is the nature of the fine-tuning dataset used in our experiment.
Due to the uncertain quality of Common Voice English, we instead used LibriSpeech, which is a carefully read speech from book texts.
Since it has been empirically known in ASR evaluation that audio samples of noisy speech can degrade performance \cite{babu2021xlsr}, the clear articulation and recording of LibriSpeech could help the model achieve better results.

\subsection{Revisiting logographicity}
The concept of logographicity used in this study measures the degree of one-to-many mapping between phonemes (pronunciation) and graphemes (orthography), following \citet{sproat-gutkin-2021-taxonomy}.
Since the task of ASR is to map pronounced words into written words, this measurement of logographicity is valid.
However, we can also think of the many-to-one mapping relationship between phonemes and graphemes as an orthographic complexity.
For example, English <gh> can be phonologically realized as either /f/ (as in <tough>), /g/ (as in <ghost>), or silent (as in <though>).
Because the attention-based metric $S_\text{token}$ used in our study only considers how much the model looks outside the target word, it is unable to look at how much the attention is spread across the target characters, failing to take this type of complexity into account.

\subsection{Broader impacts}
Understanding the factors that affect fine-tuned accuracy of the self-supervised pretrained model provides benefits to broader applications.
In an extremely low-resource setting like this study, a fine-tuned model would learn faster and better with a phonographic (\textit{i.e.}, spelled as is pronounced) writing system or phonemic transcription than with a logographic writing system.
This strategy can apply to low-resource languages with a larger number of graphemes such as Yi syllabary of the Nuosu language, the Cherokee syllabary of Cherokee, and Canadian Aboriginal syllabics of Canadian indigenous languages, providing a path to inclusion of these languages in language technologies.

Furthermore, understanding the model's flexibility of adapting to different phonological systems and struggle in learning complex writing systems confirms the strength and weakness of the self-supervised approach of pretraining.
Namely, a multilingual Wav2Vec 2.0 model is good at learning phonology of any language but not so good at writing.
This finding possibly sheds light on computational modeling of phonology and written language acquisition.
\section{Conclusion}
This study investigated what linguistic factors can confuse ASR performance of fine-tuned self-supervised models (Wav2Vec2-XLSR-53) focusing on orthographic and phonological complexities.
The experiments trained a fine-tuned model for each language and writing system, covering 25 languages and 11 writing systems in total.
The results demonstrated that speech recognition accuracy, in particular CER, strongly correlates with orthographic complexities, that is, with the size of the grapheme inventory and the degree of logographicity of a language's writing system.
On the other hand, the results showed that CER has no significant correlation with the size of the phonological inventory of the target language.
In addition, more complex orthographies turned out to make the model learn less accurately, more slowly, and less stably than phonographic writing systems.
These results confirm the robustness of the self-supervised pretrained model fine-tuned on languages with unseen phonology and the negative effect of orthographic complexities on ASR performance.
\section{Limitations}
It is worth mentioning that there are still several methodological limitations in this study.

\paragraph{Quality of the dataset.}
Since Common Voice is a dataset with speech samples collected through crowdsourcing, the quality of data may vary.
While there is a validation system to filter out poor samples, there are often speakers' mistakes and other errors in the dataset.

\paragraph{Evaluation metrics.}
Though CER is one of the most commonly used metrics in ASR, its cross-lingual applicability is questionable.
Since a letter in different orthographies can encode different lengths of phonemes, an orthography that represents multiple phonemes like Chinese characters might be more prone to errors than alphabetic orthographies that have one-to-one mapping to pronunciation.
In addition, as mentioned in Section \ref{sec:discussion}, the attention-based logographicity measure $S_\text{token}$ captures the one-to-many mapping between pronunciation and writing but not the many-to-one mapping.

\paragraph{Other Wav2Vec models.}
As mentioned in Section \ref{sec:related}, other multilingual pre-trained Wav2Vec2 models have been developed with more pretraining data and with various model parameter sizes.
We have not shown the results from these models here, and the findings in this paper do not necessarily promise reproducibility with these pretrained models.

\paragraph{Low-resource setting.}
For the sake of controlled experiments, this study limited the amount of training data for each language to be 10k seconds in total.
As described in Section \ref{sec:results}, different languages and writing systems can exhibit different learning curves, and training on a larger dataset or with more epochs might yield different performance results.

\section*{Acknowledgements}
This material is based upon work supported by the National Science Foundation under Grant No.~BCS-2109709.
\bibliography{main}

\begin{thebibliography}{22}
\expandafter\ifx\csname natexlab\endcsname\relax\def\natexlab#1{#1}\fi

\bibitem[{Ardila et~al.(2020)Ardila, Branson, Davis, Kohler, Meyer, Henretty, Morais, Saunders, Tyers, and Weber}]{ardila-etal-2020-common}
Rosana Ardila, Megan Branson, Kelly Davis, Michael Kohler, Josh Meyer, Michael Henretty, Reuben Morais, Lindsay Saunders, Francis Tyers, and Gregor Weber. 2020.
\newblock \href {https://aclanthology.org/2020.lrec-1.520} {{C}ommon {V}oice: A massively-multilingual speech corpus}.
\newblock In \emph{Proceedings of the Twelfth Language Resources and Evaluation Conference (LREC)}, pages 4218--4222.

\bibitem[{Babu et~al.(2021)Babu, Wang, Tjandra, Lakhotia, Xu, Goyal, Singh, von Platen, Saraf, Pino, Baevski, Conneau, and Auli}]{babu2021xlsr}
Arun Babu, Changhan Wang, Andros Tjandra, Kushal Lakhotia, Qiantong Xu, Naman Goyal, Kritika Singh, Patrick von Platen, Yatharth Saraf, Juan Pino, Alexei Baevski, Alexis Conneau, and Michael Auli. 2021.
\newblock \href {http://arxiv.org/abs/2111.09296} {{XLS-R}: Self-supervised cross-lingual speech representation learning at scale}.

\bibitem[{Baevski et~al.(2020)Baevski, Zhou, Mohamed, and Auli}]{baevski2020wav2vec}
Alexei Baevski, Yuhao Zhou, Abdelrahman Mohamed, and Michael Auli. 2020.
\newblock \href {https://proceedings.neurips.cc/paper/2020/hash/92d1e1eb1cd6f9fba3227870bb6d7f07-Abstract.html} {{wav2vec} 2.0: A framework for self-supervised learning of speech representations}.
\newblock In \emph{Advances in Neural Information Processing Systems (NeurIPS)}, volume~30, pages 12449--12460.

\bibitem[{Conneau et~al.(2021)Conneau, Baevski, Collobert, Mohamed, and Auli}]{conneau2020unsupervised}
Alexis Conneau, Alexei Baevski, Ronan Collobert, Abdelrahman Mohamed, and Michael Auli. 2021.
\newblock \href {https://doi.org/10.21437/Interspeech.2021-329} {Unsupervised cross-lingual representation learning for speech recognition}.
\newblock In \emph{Proceedings of INTERSPEECH}, pages 2426--2430.

\bibitem[{Coupé et~al.(2019)Coupé, Oh, Dediu, and Pellegrino}]{coupe-2019}
Christophe Coupé, Yoon~Mi Oh, Dan Dediu, and François Pellegrino. 2019.
\newblock \href {https://doi.org/10.1126/sciadv.aaw2594} {Different languages, similar encoding efficiency: Comparable information rates across the human communicative niche}.
\newblock \emph{Science Advances}, 5(9):eaaw2594.

\bibitem[{Deng et~al.(2014)Deng, Allahverdyan, Li, and Wang}]{deng-et-al-2014-rank}
Weibing Deng, Armen~E. Allahverdyan, Bo~Li, and Qiuping~A. Wang. 2014.
\newblock \href {https://doi.org/https://doi.org/10.1140/epjb/e2014-40805-2} {Rank-frequency relation for {C}hinese characters}.
\newblock \emph{The European Physical Journal~B}, 87(47).

\bibitem[{Devlin et~al.(2019)Devlin, Chang, Lee, and Toutanova}]{devlin2019bert}
Jacob Devlin, Ming-Wei Chang, Kenton Lee, and Kristina Toutanova. 2019.
\newblock \href {https://doi.org/10.18653/v1/N19-1423} {{BERT}: Pre-training of deep bidirectional {T}ransformers for language understanding}.
\newblock In \emph{Proceedings of the Conference of the North {A}merican Chapter of the Association for Computational Linguistics: Human Language Technologies (NAACL HLT)}, pages 4171--4186.

\bibitem[{Graves et~al.(2006)Graves, Fern\'{a}ndez, Gomez, and Schmidhuber}]{graves-etal-2006-ctc}
Alex Graves, Santiago Fern\'{a}ndez, Faustino Gomez, and J\"{u}rgen Schmidhuber. 2006.
\newblock \href {https://doi.org/10.1145/1143844.1143891} {Connectionist temporal classification: labelling unsegmented sequence data with recurrent neural networks}.
\newblock In \emph{Proceedings of the 23rd International Conference on Machine Learning (ICML)}, pages 369--376.

\bibitem[{Jégou et~al.(2011)Jégou, Douze, and Schmid}]{jegou2011product}
Herve Jégou, Matthijs Douze, and Cordelia Schmid. 2011.
\newblock \href {https://doi.org/10.1109/TPAMI.2010.57} {Product quantization for nearest neighbor search}.
\newblock \emph{IEEE Transactions on Pattern Analysis and Machine Intelligence}, 33(1):117--128.

\bibitem[{Koehn(2005)}]{koehn-2005-europarl}
Philipp Koehn. 2005.
\newblock \href {https://aclanthology.org/2005.mtsummit-papers.11} {{E}uroparl: A parallel corpus for statistical machine translation}.
\newblock In \emph{Proceedings of Machine Translation Summit X}, pages 79--86.

\bibitem[{Maruyama and Yamamoto(2018)}]{maruyama-yamamoto-2018-simplified}
Takumi Maruyama and Kazuhide Yamamoto. 2018.
\newblock \href {https://www.aclweb.org/anthology/L18-1185} {Simplified corpus with core vocabulary}.
\newblock In \emph{Proceedings of the Eleventh International Conference on Language Resources and Evaluation ({LREC})}.

\bibitem[{Moran and McCloy(2019)}]{phoible}
Steven Moran and Daniel McCloy. 2019.
\newblock \href {http://phoible.org} {{PHOIBLE} 2.0}.
\newblock \url{http://phoible.org}.

\bibitem[{Panayotov et~al.(2015)Panayotov, Chen, Povey, and Khudanpur}]{7178964}
Vassil Panayotov, Guoguo Chen, Daniel Povey, and Sanjeev Khudanpur. 2015.
\newblock \href {https://doi.org/10.1109/ICASSP.2015.7178964} {{L}ibrispeech: An {ASR} corpus based on public domain audio books}.
\newblock In \emph{IEEE International Conference on Acoustics, Speech and Signal Processing (ICASSP)}, pages 5206--5210.

\bibitem[{Park et~al.(2016)Park, Hong, and Cha}]{park-etal-2016-korean}
Jungyeul Park, Jeen-Pyo Hong, and Jeong-Won Cha. 2016.
\newblock \href {https://aclanthology.org/Y16-2002} {{K}orean language resources for everyone}.
\newblock In \emph{Proceedings of the 30th Pacific Asia Conference on Language, Information and Computation (PACLIC)}, pages 49--58.

\bibitem[{Park(2019)}]{park2019g2pk}
Kyubyong Park. 2019.
\newblock \href {https://github.com/Kyubyong/g2pk} {{g2pK}}.
\newblock \url{https://github.com/Kyubyong/g2pk}.

\bibitem[{Phatthiyaphaibun et~al.(2023)Phatthiyaphaibun, Chaovavanich, Polpanumas, Suriyawongkul, Lowphansirikul, Chormai, Limkonchotiwat, Suntorntip, and Udomcharoenchaikit}]{phatthiyaphaibun-etal-2023-pythainlp}
Wannaphong Phatthiyaphaibun, Korakot Chaovavanich, Charin Polpanumas, Arthit Suriyawongkul, Lalita Lowphansirikul, Pattarawat Chormai, Peerat Limkonchotiwat, Thanathip Suntorntip, and Can Udomcharoenchaikit. 2023.
\newblock \href {https://doi.org/10.18653/v1/2023.nlposs-1.4} {{P}y{T}hai{NLP}: {T}hai natural language processing in {P}ython}.
\newblock In \emph{Proceedings of the 3rd Workshop for Natural Language Processing Open Source Software (NLP-OSS 2023)}, pages 25--36.

\bibitem[{Radford et~al.(2022)Radford, Kim, Xu, Brockman, McLeavey, and Sutskever}]{radford2022robust}
Alec Radford, Jong~Wook Kim, Tao Xu, Greg Brockman, Christine McLeavey, and Ilya Sutskever. 2022.
\newblock \href {http://arxiv.org/abs/2212.04356} {Robust speech recognition via large-scale weak supervision}.

\bibitem[{Rouditchenko et~al.(2023)Rouditchenko, Khurana, Thomas, Feris, Karlinsky, Kuehne, Harwath, Kingsbury, and Glass}]{rouditchenko23_interspeech}
Andrew Rouditchenko, Sameer Khurana, Samuel Thomas, Rogerio Feris, Leonid Karlinsky, Hilde Kuehne, David Harwath, Brian Kingsbury, and James Glass. 2023.
\newblock \href {https://doi.org/10.21437/Interspeech.2023-1061} {{Comparison of Multilingual Self-Supervised and Weakly-Supervised Speech Pre-Training for Adaptation to Unseen Languages}}.
\newblock In \emph{Proceedings of INTERSPEECH}, pages 2268--2272.

\bibitem[{{Seamless Communication} et~al.(2023){Seamless Communication}, Barrault, Chung, Meglioli, Dale, Dong, Duppenthaler, Duquenne, Ellis, Elsahar, Haaheim, Hoffman, Hwang, Inaguma, Klaiber, Kulikov, Li, Licht, Maillard, Mavlyutov, Rakotoarison, Sadagopan, Ramakrishnan, Tran, Wenzek, Yang, Ye, Evtimov, Fernandez, Gao, Hansanti, Kalbassi, Kallet, Kozhevnikov, Gonzalez, Roman, Touret, Wong, Wood, Yu, Andrews, Balioglu, Chen, Costa-jussà, Elbayad, Gong, Guzmán, Heffernan, Jain, Kao, Lee, Ma, Mourachko, Peloquin, Pino, Popuri, Ropers, Saleem, Schwenk, Sun, Tomasello, Wang, Wang, Wang, and Williamson}]{communication2023seamless}
{Seamless Communication}, Loïc Barrault, Yu-An Chung, Mariano~Coria Meglioli, David Dale, Ning Dong, Mark Duppenthaler, Paul-Ambroise Duquenne, Brian Ellis, Hady Elsahar, Justin Haaheim, John Hoffman, Min-Jae Hwang, Hirofumi Inaguma, Christopher Klaiber, Ilia Kulikov, Pengwei Li, Daniel Licht, Jean Maillard, Ruslan Mavlyutov, Alice Rakotoarison, Kaushik~Ram Sadagopan, Abinesh Ramakrishnan, Tuan Tran, Guillaume Wenzek, Yilin Yang, Ethan Ye, Ivan Evtimov, Pierre Fernandez, Cynthia Gao, Prangthip Hansanti, Elahe Kalbassi, Amanda Kallet, Artyom Kozhevnikov, Gabriel~Mejia Gonzalez, Robin~San Roman, Christophe Touret, Corinne Wong, Carleigh Wood, Bokai Yu, Pierre Andrews, Can Balioglu, Peng-Jen Chen, Marta~R. Costa-jussà, Maha Elbayad, Hongyu Gong, Francisco Guzmán, Kevin Heffernan, Somya Jain, Justine Kao, Ann Lee, Xutai Ma, Alex Mourachko, Benjamin Peloquin, Juan Pino, Sravya Popuri, Christophe Ropers, Safiyyah Saleem, Holger Schwenk, Anna Sun, Paden Tomasello, Changhan Wang, Jeff Wang, Skyler Wang, and Mary
  Williamson. 2023.
\newblock \href {http://arxiv.org/abs/2312.05187} {Seamless: Multilingual expressive and streaming speech translation}.

\bibitem[{Sproat and Gutkin(2021)}]{sproat-gutkin-2021-taxonomy}
Richard Sproat and Alexander Gutkin. 2021.
\newblock \href {https://doi.org/10.1162/coli_a_00409} {The taxonomy of writing systems: How to measure how logographic a system is}.
\newblock \emph{Computational Linguistics}, 47(3):477--528.

\bibitem[{Takaoka et~al.(2018)Takaoka, Hisamoto, Kawahara, Sakamoto, Uchida, and Matsumoto}]{takaoka-etal-2018-sudachi}
Kazuma Takaoka, Sorami Hisamoto, Noriko Kawahara, Miho Sakamoto, Yoshitaka Uchida, and Yuji Matsumoto. 2018.
\newblock \href {https://aclanthology.org/L18-1355} {{S}udachi: a {J}apanese tokenizer for business}.
\newblock In \emph{Proceedings of the Eleventh International Conference on Language Resources and Evaluation ({LREC})}.

\bibitem[{Wang et~al.(2020)Wang, Ke, Zheng, Huang, Jiang, Zhu, and Huang}]{wang2020chinese}
Yida Wang, Pei Ke, Yinhe Zheng, Kaili Huang, Yong Jiang, Xiaoyan Zhu, and Minlie Huang. 2020.
\newblock \href {https://arxiv.org/abs/2008.03946} {A large-scale {C}hinese short-text conversation dataset}.
\newblock In \emph{Proceedings of the 9th CCF International Conference on Natural Language Processing and Chinese Computing (NLPCC)}.

\end{thebibliography}

\appendix

\end{document}